\documentclass[
]{ceurart}

\begin{document}

\copyrightyear{2021}
\copyrightclause{Copyright for this paper by its authors.
  Use permitted under Creative Commons License Attribution 4.0
  International (CC BY 4.0).}

\conference{Computational Humanities Research Conference 2021}

\title{Navigating the Mise-en-Page: Interpretive Machine Learning Approaches to the Visual Layouts of Multi-Ethnic Periodicals}

\author[1]{Benjamin Charles Germain Lee$^*$}[%
orcid=0000-0002-1677-6386,
email=bcgl@cs.washington.edu,
url=https://www.bcglee.com/,
]
\address[1]{University of Washington}

\author[2]{Joshua {Ortiz Baco}$^*$}[%
orcid=0000-0002-9723-4262,
email=joshuaortizbaco@utexas.edu,
]
\address[2]{Univeristy of Texas at Austin}

\author[3]{Sarah H. Salter$^*$}[%
orcid=0000-0001-7548-0932,
email=Sarah.Salter@tamucc.edu,
url=https://www.sarahhsalter.net/,
]
\address[3]{Texas A\&M University-Corpus Christi}

\author[4]{Jim Casey$^*$}[%
orcid=0000-0001-5847-5242,
email=jccasey@psu.edu,
url=https://www.jim-casey.com/,
]
\address[4]{Pennsylvania State University}
\address{$*$ = equal contribution}

\newcommand\bcgl[1]{\textcolor{red}{\textit{BCGL: #1}}}
\newcommand\jc[1]{\textcolor{blue}{\textit{JC: #1}}}
\newcommand\job[1]{\textcolor{green}{\textit{JOB: #1}}}
\newcommand\shs[1]{\textcolor{orange}{\textit{SHS: #1}}}

\begin{abstract}
This paper presents a computational method of analysis that draws from machine learning, library science, and literary studies to map the visual layouts of multi-ethnic newspapers from the late 19th and early 20th century United States. This work departs from prior approaches to newspapers that focus on individual pieces of textual and visual content. Our method combines Chronicling America's MARC data and the Newspaper Navigator machine learning dataset to identify the visual patterns of newspaper page layouts. By analyzing high-dimensional visual similarity, we aim to better understand how editors spoke and protested through the layout of their papers.
\end{abstract}


\maketitle

\section{Introduction}

This paper presents a set of developing methods for analyzing Black, Latinx, and other ethnic newspapers in the late nineteenth and early twentieth-century United States. This historical period was marked by rising tides of racism and xenophobia. Editing a newspaper dedicated to the progress of Black, Latinx, or immigrant communities often entailed facing down forces of hate and violence.

Editors in the historical multi-ethnic press often chose to proceed circumspectly. Rather than publish bold condemnations in headlines or editorials, it might have been safer to rely on other affordances of newspapers. It was possible to convey ideas using the paper’s very mise-en-page, in between the lines (or columns) through editorial strategies such as juxtaposing different content types, riffing on a given format, or deliberately reprinting unusual or out-of-context material. In the multi-ethnic press, editors often employed these strategies to render layout as its own field of meaning-making: relations between items on the page could recast or resignify seemingly distinct components of the newspaper. This feature of multi-ethnic editorial communication makes it untenable to extract and analyze isolated textual or visual items from these newspapers. 

Computational methods make it possible to analyze the visual patterns of multi-ethnic newspapers at scale. While prior research has approached these questions through textual analysis, a pivot to computational visual analysis compensates for the frequently noisy digital surrogates of surviving multi-ethnic newspapers, too often poorly preserved. 

\begin{figure}
    \centering
    \includegraphics[width=0.7\linewidth]{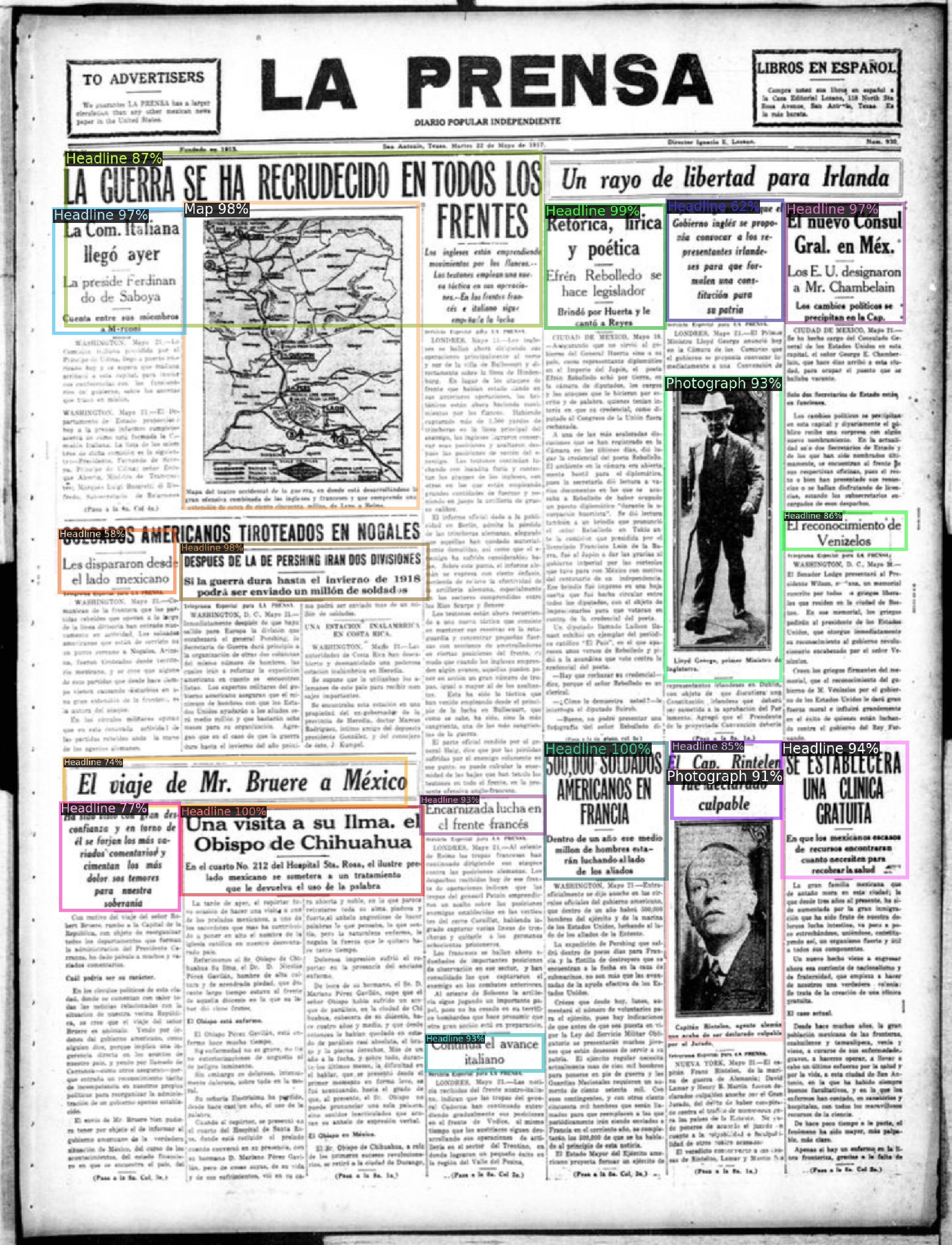}
    \caption{A visualization of the front page of the May 22, 1917, issue of \textit{La Prensa} \cite{lp_22_may_1917}, annotated with bounding boxes of visual content derived from the Newspaper Navigator dataset \cite{nn_dataset}. These bounding boxes are predictions made by a machine learning model showing, in this case, the locations of headlines, photographs, and maps (each bounding box includes the predicted class, as well as a confidence score, in the top left corner). In this paper, we utilize these bounding boxes to compare newspaper titles according to visual similarity.}
    \label{fig:la_prensa}
\end{figure}

Our working method uses machine learning to map the layouts of hundreds of thousands of newspaper pages, quantified by visual similarity, using the Newspaper Navigator dataset (as pictured in Figure \ref{fig:la_prensa}) \cite{nn_dataset}. Visualizing and comparing layouts at scale provides a wayfinding function, helping to direct archival researchers to distinctive and meaningful places in large collections of digitized newspapers and cultural heritage collections writ large.

In this paper, we describe work in progress that brings machine learning, library science, and literary studies into conversation. This work includes: (1) a dataset of visual features from 16.3 million pages of historical newspapers; (2) remediation in 207 catalog records for 309 ethnic editors; and (3) modes for reading the practical languages of editing in specific historical contexts. Prior research efforts blend together in our exploratory machine learning prototypes for quantitatively and qualitatively measuring the visual similarities of the formats of a newspaper page. These visual similarities can help illuminate the ideas and activism that flowed through the editorial craft and conduct of historical multi-ethnic newspapers. Moreover, our work formulating a ``similarity score'' affirms the promise of multi-disciplinary collaborations for improving access and analysis of larger-scale digitized newspaper collections.

\section{Related Prior Work on Developing Computational Approaches to Multi-Ethnic Periodicals}

Multiple fields are eager to explore the new possibilities for analyzing digitized collections of historical newspapers. Complementary conversations in computational, archival, and literary studies have begun to pursue those possibilities in rich but largely disparate ways. Silos persist even as many scholars make parallel use of the Library of Congress’s Chronicling America and other national newspaper digital initiatives\cite{chronam, trove, pekarek_europeana_2012, willems_europeana_2015, kreitz_american_2018, delpher, british, ourdigitalworld, paperspast}. 

This paper builds on multi-ethnic press scholarship and archival research in Chronicling America and other repositories. The prevailing view of political activism in historical multi-ethnic newspapers has tended to focus on the textual and visual content that drove demands for social justice \cite{nerone_us_2003, cole_how_2020, williams_cultivating_2015, stein_2007}. A complementary approach has focused on social histories, tracing the networks of a paper's contributors and readers \cite{foster_narrative_2005, gardner_black_2015, rhodes_mary_1998, spires_practice_2019, fagan_black_2016}. What can be easy to miss, however, is that editors can use the format of a paper itself to voice critiques and protests. The formal nature of editors' craft is difficult to grasp in the meticulous approaches to archival research on a case-by-case basis. Our work aims to develop experimental machine learning applications to map the larger patterns of editorial practices and editing conventions that span languages, communities, and eras.

Conversations in library and archival studies are exploring how best to account for histories of race and ethnicity in metadata. Rather than being reduced to a theme or topic within a collection, specialists have rethought how race and ethnicity might work as organizing principles for collecting and describing historical materials. Race and ethnicity as categories of identity are historically contingent and socially constructed. That fluidity challenges the rigors of knowledge organization systems that impose stable categories and schemas. For some archivists, however, categories of race and identity productively serve as a form of provenance to guide collection development and access policies \cite{wurl_ethnicity_2005, daniel_archival_2014, daniel_documenting_2014}. Many of these changes at the collection level are being addressed in the metadata and, more recently, through linked open data. Because the development of controlled vocabularies, such as the Library of Congress Subject Headings, and Machine Readable Cataloging (MARC) largely predate sensitivity to these histories in libraries and archives, metadata has become a vital space for rediscovering the historical editors of multi-ethnic newspapers \cite{ventura_recovering_2019}. 

At the same time, an overabundance of work with more “computationally robust,” institutionally privileged white print collections has prompted preliminary conversations regarding the directions of computational humanities research. Indeed, from OCR engines that do not support indigenous languages or different dialects, to entire subfields of natural language processing and textual analysis that assume English as the default language of study, algorithmically-mediated erasure continues to be uncovered and detailed by scholars \cite{ fagan_chronicling_2016, alpert_abrams_machine_2016, joshi_20, yakel_archival_2003, hardy, williams_cas, gustafson_ethnic_2015, risam_new_2018, noble_2018}. Many have advocated for the development of computational approaches for those collections that remain marginal. This paper concerns one such corrective for digitized periodicals: utilizing machine learning to study visual content and page layout rather than textual content, circumventing the use of OCR engines entirely.

Computational scholars have isolated visual content within periodicals in order to produce datasets of isolated content. Humanistic scholars have utilized these atomized items for interpretation. We are instead interested in the visual layouts themselves and the relations they imply on the page. By treating relations and patterns within layouts as forms of interpretive content -- reading them as a ``visual grammar'' of photographs, advertisements, headlines, and beyond -- we are in effect exploring a publication’s creative history, local contexts, and circulation networks.

\section{Methods}

The work described in this paper began in part with the Newspaper Navigator project, created as part of Library of Congress’s Innovator in Residence program in 2019 with the goal of excavating visual content in Chronicling America. Developed in partnership with LC Labs, the National Digital Newspaper Program, and IT Design \& Development at the Library of Congress, Newspaper Navigator utilizes a machine learning model to extract seven visual features from 16 million pages in Chronicling America: advertisements, photographs, illustrations, maps, comics, editorial cartoons, and headlines \cite{nn_dataset}. This machine learning model consists of an object detection model that had been finetuned on thousands of bounding box annotations of visual content created by volunteers as part of the Beyond Words crowdsourcing initiative launched by LC Labs in 2017 \cite{beyond_words}. The resulting extracted visual content was publicly released as the Newspaper Navigator dataset, available in the public domain for all to use \cite{nn_dataset}.

Newspaper Navigator coincides with current trends in periodical studies. As Sean Latham and Robert Scholes argue in an influential essay, ``We have often been too quick to see magazines [and other periodicals] as containers of discrete bits of information rather than autonomous objects of study'' \cite{latham_scholes_2006}. In practice, this means that periodical scholars tend to examine not only the individual pieces of content in a given publication, but also its overall construction, taking into account such architectural elements as a publication’s formats, columns, and visual layout. The typical methods to apply this approach require intensive reading of individual newspaper pages. Researchers gradually develop an intuition about the general style and flavor of a certain publication. A publication's implicit patterns can be relatively intangible because periodical formats are necessarily fixed—to preserve continuity—and flexible—to accommodate new and changing scenarios. A paper looks like itself almost always, but not always. Even beyond academic inquiry, dedicated readers of any periodical often develop this kind of intuitive mental map of a publication’s page designs and formats. 

We saw the possibility of combining these two types of reading. On one hand, Newspaper Navigator makes it possible to chart visual features across the entirety of Chronicling America. On the other hand, the methods in periodical studies encourage attention to the juxtaposition of visual features on a page. Starting with these two methods, we developed a set of preliminary working questions. How do the visual features on the page of a newspaper interlock and mutually inform each other? If we look at all of those features holistically, what kinds of styles and ideas can we detect over the lifespan of a single newspaper? More broadly, what added perspective can we gain by using the Newspaper Navigator to trace the sum of newspaper layouts over hundreds of publications? Finally, can we establish a “baseline” for newspaper production in the later 19th and early 20th century United States that would allow us to see the critical departures made by multi-ethnic newspapers in the service of social justice during moments of elevated racism or xenophobia? 

\subsection{Methods I: Adapting MARC 21 Metadata for Research}

Adapting the Newspaper Navigator dataset to answer these questions required us to collect additional data about the subsets of multi-ethnic newspapers within Chronicling America. This information was partially extracted from the bibliographic data of newspaper titles in the machine-readable cataloging (MARC) format. All records for newspapers in Chronicling America follow the cataloging standards of the Cooperative ONline SERials Program (CONSER) \cite{conser_about_loc}, providing a somewhat normalized set of features for sorting newspapers. Additionally, Chronicling America’s partnering institutions provide a “title essay” that describes and contextualizes each newspaper. These unstructured data has recently been the focus of efforts for metadata enhancement in remediation projects geared to identifying ethnic editors \cite{ortiz_baco_title_2019}.

The last step in collecting and organizing our data was combining MARC fields and structured data generated from the text of title essays to find relevant publications. Metadata from the title essays allowed us to select only ethnic newspapers with identifiable editors. From this subset, we selected MARC fields used by cataloguers to record title changes, race, ethnicity, and languages associated with each publication. We initially hoped to find an indicator of visual change within the metadata for newspapers. We considered that serials with more than one language edition, language notes, or different titles over their publication runs would feature changes to visual and layout elements of newspapers to mark these relationships or changes.\footnote{The 780 and 785 fields indicate all preceding and succeeding titles for a publication, respectively. The 546 field describes the language of a resource while the 765 indicates the original language of a translated publication and the 775 captures other language editions of a publication} Title changes, for example, are a cataloguing practice for linking preceding and succeeding titles of a publication that has had a substantial name change, which we assumed would also produce modifications to the graphic elements on the page. We were able to parse these metadata fields where the change inherent in the seriality of newspapers is recorded, which in turn provided parameters for capturing difference, similarity, and linking in an otherwise static knowledge organization system.

\begin{figure}
    \centering
    \includegraphics[width=1.0\linewidth]{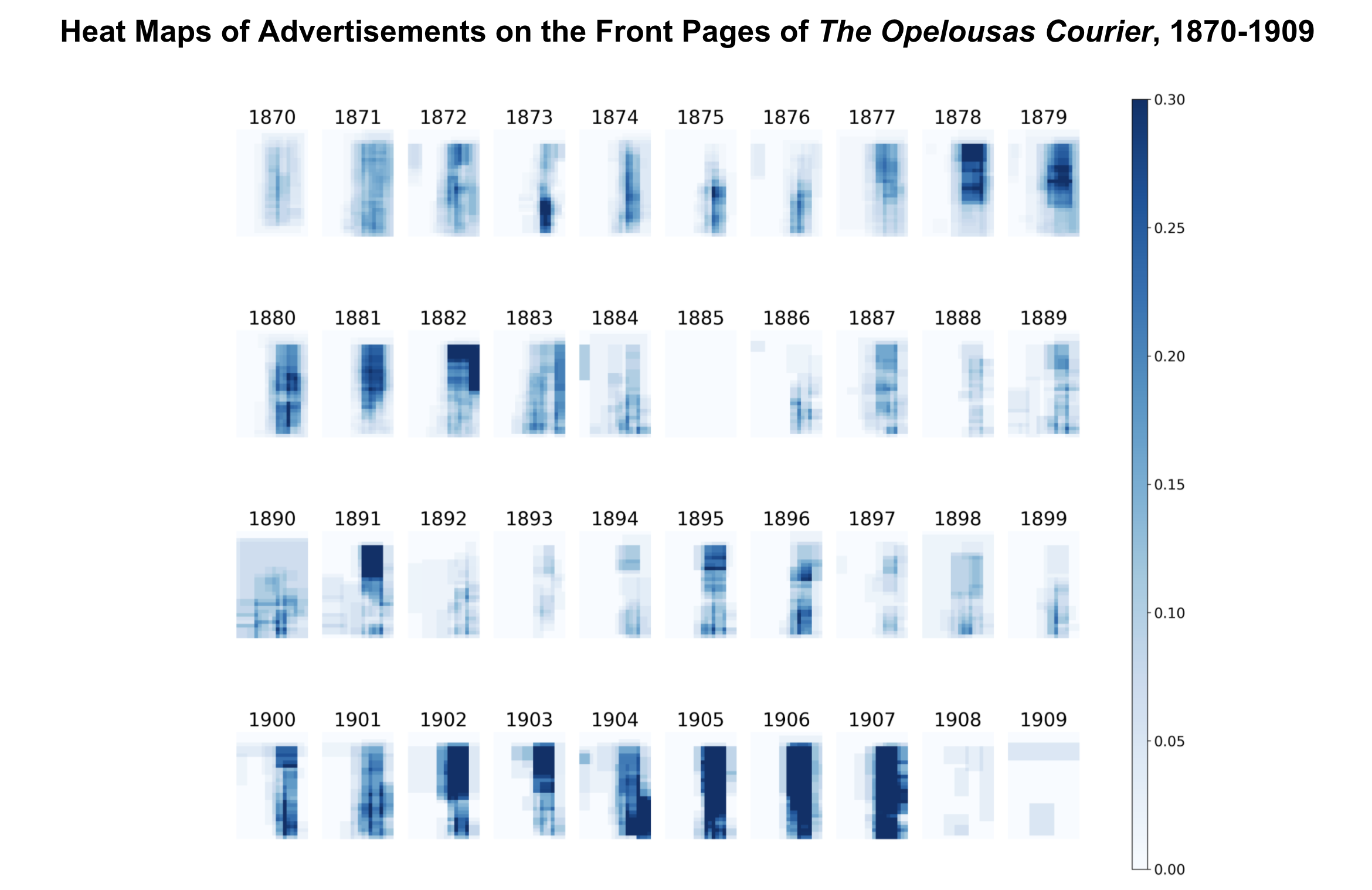}
    \caption{A heatmap of advertisements appearing on the front page of the Opelousas Courier, 1870-1909. Darker regions on the heatmaps correspond to a higher concentration of advertisement pages appearing in that region in aggregate over a given year.}
    \label{fig:opelousas_heat_maps}
\end{figure}

\subsection{Methods 2: Quantifying and Visualizing the Mise-en-Page}

The first visual output we experimented with were heat maps, created with Newspaper Navigator data that encodes the positions of visual features on a newspaper page. Inspired by PageOneX \cite{pageonex}, these heat maps concisely summarize visual content patterns in an interpretable fashion. Figure \ref{fig:opelousas_heat_maps} shows the locations and relationships of advertisements across four decades of \textit{The Opelousas Courier},\footnote{\textit{The Opelousas Courier} is described in Chronicling America as a Louisiana multilingual newspaper, with content in French and English. \url{https://chroniclingamerica.loc.gov/lccn/sn83026389/marc/}.
} As an initial experiment in visual layout representation, the heat maps enabled us to trace the development of discrete visual features, otherwise impossible to quantify by examining front pages by eye.

This approach facilitates several comparative opportunities. We rendered multiple visual categories into heat maps to understand the interplay between visual components in a single publication. 
We also utilized the heat map graphics to compare different newspapers using the same visual categories. The heat maps offered a useful preliminary system for studying newspaper layout as historical information. 

Heat maps added greater nuance to our historical questions about newspaper layout. For example, what are the relationships between visual content, editorial practices, and print technology? What are the roles of regional and local context in predicting layout? How do we account for journalistic trends, such as the rise of syndicated content or nationally-circulating advertisements?

The heat maps showed the need for further exploration. In analyzing one publication at a time, the heat maps reproduced the methods of archival periodical studies, even in visually abstracted ways. While the MARC records helped us to identify the titles we were most interested in rendering into heat maps, this piecemeal approach to newspaper metadata and layouts proved only a first step. 

\subsection{Methods 3: Toward a Visual Constellation of Multi-Ethnic Newspapers}

Our next experiment attempted to scale up possible representations to refocus on illustrating change and similarity across tens, hundreds, and even thousands of publications. Our resulting exploratory model utilized the quantified heat maps to define a metric over visual similarity. In particular, the distance between two newspaper titles can be captured by the residuals from subtracting two heatmaps from one another. As an initial approach, we segmented each newspaper title’s run into individual years. For each title and year, we then treated the seven resulting heat maps (one for advertisements, one for headlines, etc.) collectively as a single, high-dimensional vector. In order to interpret the clusters in this high-dimensional space, we utilized T-SNE for dimensionality reduction \cite{tsne}.

\begin{figure}
    \centering
    \includegraphics[width=1.0\linewidth]{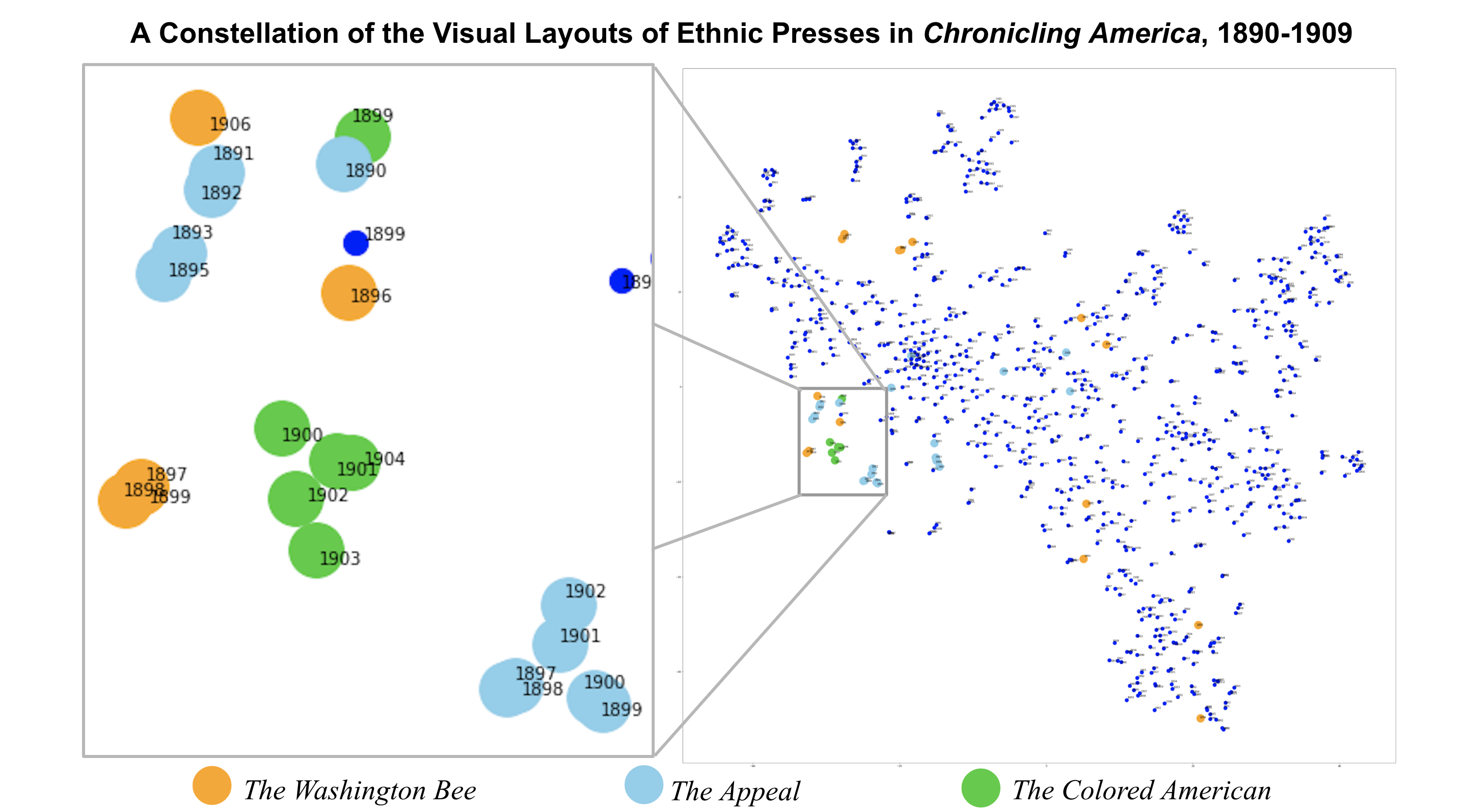}
    \caption{A 2-dimensional map of visual similarity across all ethnic titles in Chronicling America, filtered using \cite{ortiz_baco_title_2019}. Each point on the similarity map represents the composite front page of a given newspaper title for a given year, from 1890 to 1909 (individual years are labeled on the visualization). The magnified cluster reveals that \textit{The Washington Bee}, \textit{The Appeal}, and \textit{The Colored American} are largely grouped together across years (individual front pages from each title are shown in Figure \ref{fig:sample_front_pages}). Notably, all three titles are from the Black press.}
    \label{fig:constellation}
\end{figure}

\begin{figure}
    \centering
    \includegraphics[width=1.0\linewidth]{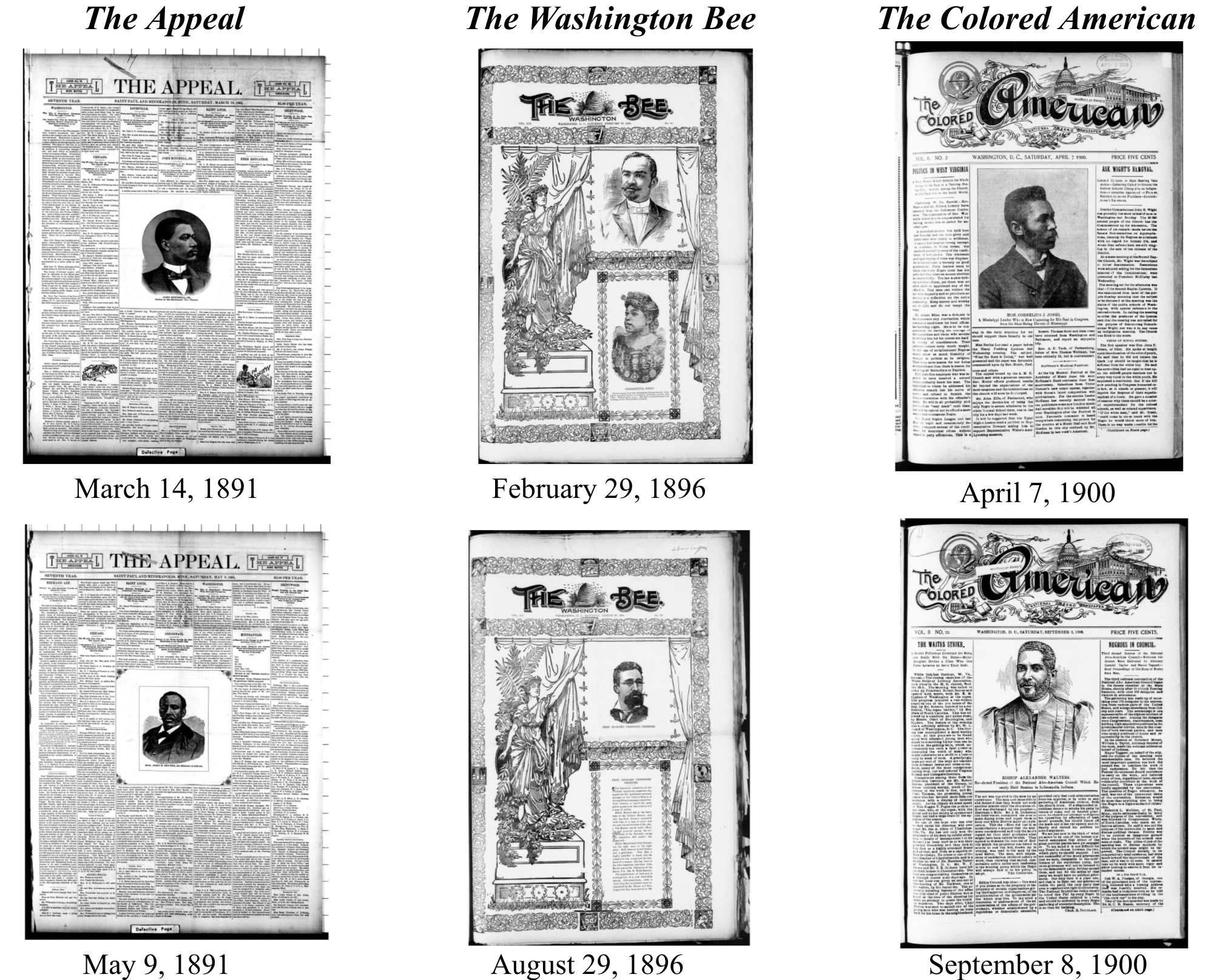}
    \caption{Sample front pages from \textit{The Appeal}, \textit{The Washington Bee}, and \textit{The Colored American} from 1891, 1896, and 1900, respectively -- all of which are title-year pairs appearing in the same cluster in Figure \ref{fig:constellation} \cite{appeal_14_march_1891, appeal_9_may_1891, bee_29_feb_1896, bee_29_aug_1896, ca_29_feb_1900, ca_8_sep_1900}. All six front pages feature visual content -- namely, illustrations and photographs of individuals -- prominently in the center of the page.}
    \label{fig:sample_front_pages}
\end{figure}

This process resulted in exploratory constellations of newspaper titles, as shown in Figure \ref{fig:constellation}. In this figure, we also show a close-up of one such ``similarity cluster,'' revealing that three Black newspapers (\textit{The Washington Bee}, \textit{The Appeal}, and \textit{The Colored American}) are largely grouped together across years. In Figure \ref{fig:sample_front_pages}, we show example front pages for these three titles from years included within the cluster. All six pictured front pages feature visual content on the center of the page. Notably, all six feature photographs and illustrations of individuals, a trend that is even more apparent when browsing these front pages across different issues. These images demonstrate that the similarity clusters, generated using Newspaper Navigator’s rubric of seven content classes, accurately reflect common visual patterns of layout in distinct newspapers across the Black newspapers held in Chronicling America. The shared qualities across each of the three newspapers pictured in Figure \ref{fig:sample_front_pages} are suggestive. They are roughly contemporaneous African American newspapers that locate specific types of images near the center of the front page. These results indicate the capacity of the method to produce a multifaceted research question: what can we learn from the use of visual culture and portraiture in the material pages of late nineteenth-century African American newspapers? 

These constellations of similarity provide a means to uncover small networks to focus on and interpret at a much more granular level. This is not a fully quantitative representation, but rather an exploratory framework for identifying and interpreting networks of related publications and the editorial practices that shape them. Given our overarching interest in the historical trajectories and meanings of newspaper formats, these visualizations show great promise. The constellation map is a model for how to convert raw bounding box information into interpretable provocations about historical similarity. Similarity clusters introduce visual layouts as clues to a shared context, inviting new research into the different historical traditions, technological moments, or political causes that comprise the histories of editorship.

\section{Findings \& Future Work}

Our ongoing collaboration offers three central findings to date. First, this work finds that full level CONSER MARC records provide useful data features in machine-learning applications. Specifically, the most value comes from fields that register change over time, such as preceding and succeeding titles, and describe relationships between different editions of the same publication. Further metadata enhancement focused on identifying ethnic editors will significantly improve the interpretative value of visualizations and pattern recognition from large-scale collections like Chronicling America. 

Second, this collaboration reveals how computational notions of layout similarity can be leveraged for humanistic inquiry. Using the machine learning-constructed Newspaper Navigator dataset of visual content, we have generated heat maps and constellation maps for exploring trends of visual layout within and across newspaper titles, respectively. These visualizations provide us with new visual grammars and affordances for excavating editorial practices. 

Third, this work required us to find a collaborative working process for creating multidisciplinary methods of analysis that engage with machine learning, library science, and the humanities. Rather than trying to blend or transplant our respective research methods, we focused on creating a shared vocabulary that could speak to current research agendas in each of the relevant fields. This set of theoretical exchanges required iterative conversations, frequently pedagogical in nature, to find a method that is not merely extractive and static but generative and dynamic. This process will help guide the future growth of conversations in computational periodicals research.

Our roadmap for future work is guided by the provocations offered throughout this paper. First, we plan to further utilize MARC data and the constellation maps in concert with one another in order to inform our understanding of both. The MARC data has the capacity to guide our navigation and understanding of the constellation maps; conversely, the constellation maps provide mechanisms for record enhancement by allowing scholars, catalogers, and librarians alike to sift through incomplete legacy records of ethnic publications. We offer this direction of future work in pursuit of uncovering editorship in ethnic presses. We plan to refine our similarity metrics and visualizations from a computational perspective in pursuit of this goal as well. The results of this research will drive new inquiry into the longer and varied histories of radical dissent and veiled protest in the multi-ethnic press. 

\begin{acknowledgments}
The authors would like to thank and acknowledge Dr. Molly Hardy, Senior Program Officer at the National Endowment for the Humanities, along with Georgia Higley and Vanessa Mitchell from the Serial and Government Publications Division at the Library of Congress, for sharing their expertise. This work would not be possible without Dr. Hardy’s input and advice.
\end{acknowledgments}

%%
%% Define the bibliography file to be used
\bibliography{main}

\begin{thebibliography}{45}
\expandafter\ifx\csname natexlab\endcsname\relax\def\natexlab#1{#1}\fi
\providecommand{\url}[1]{\texttt{#1}}
\providecommand{\href}[2]{#2}
\providecommand{\path}[1]{#1}
\providecommand{\DOIprefix}{doi:}
\providecommand{\ArXivprefix}{arXiv:}
\providecommand{\URLprefix}{URL: }
\providecommand{\Pubmedprefix}{pmid:}
\providecommand{\doi}[1]{\href{http://dx.doi.org/#1}{\path{#1}}}
\providecommand{\Pubmed}[1]{\href{pmid:#1}{\path{#1}}}
\providecommand{\bibinfo}[2]{#2}
\ifx\xfnm\relax \def\xfnm[#1]{\unskip,\space#1}\fi
%Type = Misc
\bibitem[{of~Congress(1917)}]{lp_22_may_1917}
\bibinfo{author}{L.~of~Congress}, \bibinfo{title}{La prensa, 22 may 1917.
  chronicling america: Historic american newspapers.}, \bibinfo{year}{1917}.
  \URLprefix
  \url{https://chroniclingamerica.loc.gov/lccn/sn83045395/1917-05-22/ed-1/seq-1/}.
%Type = Inbook
\bibitem[{Lee et~al.(2020)Lee, Mears, Jakeway, Ferriter, Adams, Yarasavage,
  Thomas, Zwaard, and Weld}]{nn_dataset}
\bibinfo{author}{B.~C.~G. Lee}, \bibinfo{author}{J.~Mears},
  \bibinfo{author}{E.~Jakeway}, \bibinfo{author}{M.~Ferriter},
  \bibinfo{author}{C.~Adams}, \bibinfo{author}{N.~Yarasavage},
  \bibinfo{author}{D.~Thomas}, \bibinfo{author}{K.~Zwaard},
  \bibinfo{author}{D.~S. Weld}, \bibinfo{title}{The Newspaper Navigator
  Dataset: Extracting Headlines and Visual Content from 16 Million Historic
  Newspaper Pages in Chronicling America}, \bibinfo{publisher}{Association for
  Computing Machinery}, \bibinfo{address}{New York, NY, USA},
  \bibinfo{year}{2020}, p. \bibinfo{pages}{3055–3062}. \URLprefix
  \url{https://doi-org.offcampus.lib.washington.edu/10.1145/3340531.3412767}.
%Type = Misc
\bibitem[{of~Congress \& National Endowment for~the Humanities(2021)}]{chronam}
\bibinfo{author}{L.~of~Congress \& National Endowment for~the Humanities},
  \bibinfo{title}{Chronicling {America}}, \bibinfo{year}{2021}. \URLprefix
  \url{https://chroniclingamerica.loc.gov/}.
%Type = Inproceedings
\bibitem[{Cassidy(2016)}]{trove}
\bibinfo{author}{S.~Cassidy},
\newblock \bibinfo{title}{Publishing the trove newspaper corpus},
\newblock in: \bibinfo{booktitle}{Proceedings of the Tenth International
  Conference on Language Resources and Evaluation ({LREC}'16)},
  \bibinfo{publisher}{European Language Resources Association (ELRA)},
  \bibinfo{address}{Portoro{\v{z}}, Slovenia}, \bibinfo{year}{2016}, pp.
  \bibinfo{pages}{4520--4525}. \URLprefix
  \url{https://www.aclweb.org/anthology/L16-1715}.
%Type = Inproceedings
\bibitem[{Pekárek and Willems(2012)}]{pekarek_europeana_2012}
\bibinfo{author}{A.~Pekárek}, \bibinfo{author}{M.~Willems},
\newblock \bibinfo{title}{The {Europeana} {Newspapers} – {A} {Gateway} to
  {European} {Newspapers} {Online}},
\newblock in: \bibinfo{editor}{M.~Ioannides}, \bibinfo{editor}{D.~Fritsch},
  \bibinfo{editor}{J.~Leissner}, \bibinfo{editor}{R.~Davies},
  \bibinfo{editor}{F.~Remondino}, \bibinfo{editor}{R.~Caffo} (Eds.),
  \bibinfo{booktitle}{Progress in {Cultural} {Heritage} {Preservation}},
  \bibinfo{publisher}{Springer Berlin Heidelberg}, \bibinfo{address}{Berlin,
  Heidelberg}, \bibinfo{year}{2012}, pp. \bibinfo{pages}{654--659}.
%Type = Article
\bibitem[{Willems and Atanassova(2015)}]{willems_europeana_2015}
\bibinfo{author}{M.~Willems}, \bibinfo{author}{R.~Atanassova},
\newblock \bibinfo{title}{Europeana {Newspapers}: searching digitized
  historical newspapers from 23 {European} countries},
\newblock \bibinfo{journal}{Insights} \bibinfo{volume}{28}
  (\bibinfo{year}{2015}) \bibinfo{pages}{51--56}. \URLprefix
  \url{http://insights.uksg.org/articles/10.1629/uksg.218/}.
  \DOIprefix\doi{10.1629/uksg.218}, \bibinfo{note}{number: 1 Publisher: UKSG in
  association with Ubiquity Press}.
%Type = Article
\bibitem[{Kreitz(2018)}]{kreitz_american_2018}
\bibinfo{author}{K.~Kreitz},
\newblock \bibinfo{title}{American {Alternatives}: {Participatory} {Futures} of
  {Print} from {New} {York} {City}’s {Nineteenth}-{Century}
  {Spanish}-{Language} {Press}},
\newblock \bibinfo{journal}{American Literary History} \bibinfo{volume}{30}
  (\bibinfo{year}{2018}) \bibinfo{pages}{677--702}.
  \DOIprefix\doi{https://doi.org/10.1093/alh/ajy032}.
%Type = Misc
\bibitem[{del(NA)}]{delpher}
\bibinfo{title}{About delpher}, \bibinfo{year}{N/A}. \URLprefix
  \url{https://www.delpher.nl/nl/platform/pages/helpitems?title=wat+is+delpher}.
%Type = Misc
\bibitem[{bri(NA)}]{british}
\bibinfo{title}{About the british newspaper archive}, \bibinfo{year}{N/A}.
  \URLprefix \url{https://www.britishnewspaperarchive.co.uk/help/about}.
%Type = Misc
\bibitem[{our(NA)}]{ourdigitalworld}
\bibinfo{title}{Ourdigitalworld: Digital newspapers}, \bibinfo{year}{N/A}.
  \URLprefix \url{https://ourdigitalworld.net/what-we-do/digital-newspapers/}.
%Type = Misc
\bibitem[{pap(NA)}]{paperspast}
\bibinfo{title}{Papers past},
  \bibinfo{howpublished}{\url{https://natlib.govt.nz/collections/a-z/papers-past}},
  \bibinfo{year}{N/A}.
%Type = Article
\bibitem[{Nerone and Barnhurst(2003)}]{nerone_us_2003}
\bibinfo{author}{J.~Nerone}, \bibinfo{author}{K.~G. Barnhurst},
\newblock \bibinfo{title}{{US} newspaper types, the newsroom, and the division
  of labor, 1750–2000},
\newblock \bibinfo{journal}{Journalism Studies} \bibinfo{volume}{4}
  (\bibinfo{year}{2003}) \bibinfo{pages}{435--449}. \URLprefix
  \url{https://doi.org/10.1080/1461670032000136541}.
  \DOIprefix\doi{10.1080/1461670032000136541}, \bibinfo{note}{publisher:
  Routledge \_eprint: https://doi.org/10.1080/1461670032000136541}.
%Type = Book
\bibitem[{Cole(2020)}]{cole_how_2020}
\bibinfo{author}{J.~L. Cole}, \bibinfo{title}{How the {Other} {Half} {Laughs}:
  {The} {Comic} {Sensibility} in {American} {Culture}, 1895-1920},
  \bibinfo{publisher}{U Press of Mississippi}, \bibinfo{address}{Jackson},
  \bibinfo{year}{2020}.
%Type = Article
\bibitem[{Williams(2015)}]{williams_cultivating_2015}
\bibinfo{author}{A.~N. Williams},
\newblock \bibinfo{title}{Cultivating {Black} {Visuality}: {The} {Controversy}
  over {Cartoons} in the {Indianapolis} {Freeman}},
\newblock \bibinfo{journal}{American Periodicals: A Journal of History \&
  Criticism} \bibinfo{volume}{25} (\bibinfo{year}{2015})
  \bibinfo{pages}{124--138}. \URLprefix
  \url{http://muse.jhu.edu/article/601461}.
  \DOIprefix\doi{10.1353/amp.2015.0032}.
%Type = Article
\bibitem[{Stein(2007)}]{stein_2007}
\bibinfo{author}{S.~A. Stein},
\newblock \bibinfo{title}{Making jews modern: The yiddish and ladino press in
  the russian and ottoman empires, the modern jewish experience},
\newblock \bibinfo{journal}{International Journal of Middle East Studies}
  \bibinfo{volume}{41} (\bibinfo{year}{2007}).
  \DOIprefix\doi{10.1017/S0020743809090874}.
%Type = Article
\bibitem[{Foster(2005)}]{foster_narrative_2005}
\bibinfo{author}{F.~S. Foster},
\newblock \bibinfo{title}{A {Narrative} of the {Interesting} {Origins} and
  ({Somewhat}) {Surprising} {Developments} of {African}-{American} {Print}
  {Culture}},
\newblock \bibinfo{journal}{American Literary History} \bibinfo{volume}{17}
  (\bibinfo{year}{2005}) \bibinfo{pages}{714--740}. \URLprefix
  \url{http://muse.jhu.edu/journals/american_literary_history/v017/17.4foster.html},
  \bibinfo{note}{{\textless}p{\textgreater}Volume 17, Number 4, Winter
  2005{\textless}/p{\textgreater}}.
%Type = Book
\bibitem[{Gardner(2015)}]{gardner_black_2015}
\bibinfo{author}{E.~Gardner}, \bibinfo{title}{Black {Print} {Unbound}: {The}
  {Christian} {Recorder}, {African} {American} {Literature}, and {Periodical}
  {Culture}}, \bibinfo{publisher}{Oxford University Press},
  \bibinfo{year}{2015}. \bibinfo{note}{Google-Books-ID: 76n2CQAAQBAJ}.
%Type = Book
\bibitem[{Rhodes(1998)}]{rhodes_mary_1998}
\bibinfo{author}{J.~Rhodes}, \bibinfo{title}{Mary {Ann} {Shadd} {Cary}: the
  {Black} {Press} and {Protest} in the {Nineteenth} {Century}},
  \bibinfo{publisher}{Indiana University Press}, \bibinfo{address}{Bloomington,
  IN}, \bibinfo{year}{1998}.
%Type = Book
\bibitem[{Spires(2019)}]{spires_practice_2019}
\bibinfo{author}{D.~R. Spires}, \bibinfo{title}{The practice of citizenship:
  {Black} politics and print culture in the early {United} {States}},
  \bibinfo{publisher}{University of Pennsylvania Press},
  \bibinfo{address}{Philadelphia}, \bibinfo{year}{2019}.
%Type = Book
\bibitem[{Fagan(2016)}]{fagan_black_2016}
\bibinfo{author}{B.~Fagan}, \bibinfo{title}{The {Black} {Newspaper} and the
  {Chosen} {Nation}}, \bibinfo{publisher}{University of Georgia Press},
  \bibinfo{address}{Athens, GA}, \bibinfo{year}{2016}.
%Type = Article
\bibitem[{Wurl(2005)}]{wurl_ethnicity_2005}
\bibinfo{author}{J.~Wurl},
\newblock \bibinfo{title}{Ethnicity as {Provenance}: in {Search} of {Values}
  and {Principles} for {Documenting} the {Immigrant} {Experience}},
\newblock \bibinfo{journal}{Archival Issues} \bibinfo{volume}{29}
  (\bibinfo{year}{2005}) \bibinfo{pages}{65--76}. \URLprefix
  \url{http://www.jstor.org/stable/41102095}, \bibinfo{note}{publisher: Midwest
  Archives Conference}.
%Type = Article
\bibitem[{Daniel(2014)}]{daniel_archival_2014}
\bibinfo{author}{D.~Daniel},
\newblock \bibinfo{title}{Archival representations of immigration and ethnicity
  in {North} {American} history: from the ethnicization of archives to the
  archivization of ethnicity},
\newblock \bibinfo{journal}{Archival Science} \bibinfo{volume}{14}
  (\bibinfo{year}{2014}) \bibinfo{pages}{169--203}. \URLprefix
  \url{http://link.springer.com/10.1007/s10502-013-9209-6}.
  \DOIprefix\doi{10.1007/s10502-013-9209-6}.
%Type = Incollection
\bibitem[{Bastian(2014)}]{daniel_documenting_2014}
\bibinfo{author}{J.~A. Bastian},
\newblock \bibinfo{title}{Documenting {Communities} {Through} the {Lens} of
  {Collective} {Memory}},
\newblock in: \bibinfo{editor}{D.~Daniel}, \bibinfo{editor}{A.~S. Levi} (Eds.),
  \bibinfo{booktitle}{Identity palimpsests: archiving ethnicity in the {U}.{S}.
  and {Canada}}, number \bibinfo{number}{book 5} in \bibinfo{series}{Archives,
  archivists and society}, \bibinfo{publisher}{Litwin Books},
  \bibinfo{address}{Sacramento, CA}, \bibinfo{year}{2014}, pp.
  \bibinfo{pages}{15--34}.
%Type = Article
\bibitem[{Ventura et~al.(2019)Ventura, Gauthereau, and
  Villarroel}]{ventura_recovering_2019}
\bibinfo{author}{G.~B. Ventura}, \bibinfo{author}{L.~Gauthereau},
  \bibinfo{author}{C.~Villarroel},
\newblock \bibinfo{title}{Recovering the {US} {Hispanic} {Literary} {Heritage}:
  {A} {Case} {Study} on {US} {Latina}/o {Archives} and {Digital} {Humanities}},
\newblock \bibinfo{journal}{Preservation, Digital Technology \& Culture
  (PDT\&C)} \bibinfo{volume}{48} (\bibinfo{year}{2019})
  \bibinfo{pages}{17--27}. \URLprefix
  \url{https://www.degruyter.com/document/doi/10.1515/pdtc-2018-0031/html}.
  \DOIprefix\doi{10.1515/pdtc-2018-0031}.
%Type = Article
\bibitem[{Fagan(2016)}]{fagan_chronicling_2016}
\bibinfo{author}{B.~Fagan},
\newblock \bibinfo{title}{Chronicling {White} {America}.},
\newblock \bibinfo{journal}{American Periodicals} \bibinfo{volume}{26}
  (\bibinfo{year}{2016}) \bibinfo{pages}{10--13}. \URLprefix
  \url{http://search.proquest.com/docview/2152559277/F95B44D158E64163PQ/13}.
%Type = Article
\bibitem[{Alpert-Abrams(2016)}]{alpert_abrams_machine_2016}
\bibinfo{author}{H.~Alpert-Abrams},
\newblock \bibinfo{title}{Machine {Reading} the {Primeros} {Libros}},
\newblock \bibinfo{journal}{Digital Humanities Quarterly} \bibinfo{volume}{010}
  (\bibinfo{year}{2016}).
%Type = Inproceedings
\bibitem[{Joshi et~al.(2020)Joshi, Santy, Budhiraja, Bali, and
  Choudhury}]{joshi_20}
\bibinfo{author}{P.~Joshi}, \bibinfo{author}{S.~Santy},
  \bibinfo{author}{A.~Budhiraja}, \bibinfo{author}{K.~Bali},
  \bibinfo{author}{M.~Choudhury},
\newblock \bibinfo{title}{The state and fate of linguistic diversity and
  inclusion in the {NLP} world},
\newblock in: \bibinfo{booktitle}{Proceedings of the 58th Annual Meeting of the
  Association for Computational Linguistics}, \bibinfo{publisher}{Association
  for Computational Linguistics}, \bibinfo{address}{Online},
  \bibinfo{year}{2020}, pp. \bibinfo{pages}{6282--6293}. \URLprefix
  \url{https://www.aclweb.org/anthology/2020.acl-main.560}.
  \DOIprefix\doi{10.18653/v1/2020.acl-main.560}.
%Type = Article
\bibitem[{Yakel(2003)}]{yakel_archival_2003}
\bibinfo{author}{E.~Yakel},
\newblock \bibinfo{title}{Archival representation},
\newblock \bibinfo{journal}{Archival Science} \bibinfo{volume}{3}
  (\bibinfo{year}{2003}) \bibinfo{pages}{1--25}. \URLprefix
  \url{https://doi.org/10.1007/BF02438926}. \DOIprefix\doi{10.1007/BF02438926}.
%Type = Misc
\bibitem[{Hardy and DiCuirci(2019)}]{hardy}
\bibinfo{author}{M.~Hardy}, \bibinfo{author}{L.~DiCuirci},
  \bibinfo{title}{Critical {Cataloging} and the {Serials} {Archive}: {The}
  {Digital} {Making} of “{Mill} {Girls} in {Nineteenth}-{Century} {Print}”
  - {Archive} {Journal}}, \bibinfo{year}{2019}. \URLprefix
  \url{http://www.archivejournal.net/?p=8073}.
%Type = Inproceedings
\bibitem[{Williams(2019)}]{williams_cas}
\bibinfo{author}{L.~Williams},
\newblock \bibinfo{title}{What computational archival science can learn from
  art history and material culture studies},
\newblock in: \bibinfo{booktitle}{2019 IEEE International Conference on Big
  Data (Big Data)}, \bibinfo{year}{2019}, pp. \bibinfo{pages}{3153--3155}.
  \DOIprefix\doi{10.1109/BigData47090.2019.9006527}.
%Type = Article
\bibitem[{Gustafson(2015)}]{gustafson_ethnic_2015}
\bibinfo{author}{K.~L. Gustafson},
\newblock \bibinfo{title}{Ethnic newspaper producers face archiving
  challenges},
\newblock \bibinfo{journal}{Newspaper Research Journal} \bibinfo{volume}{36}
  (\bibinfo{year}{2015}) \bibinfo{pages}{314--327}. \URLprefix
  \url{https://doi.org/10.1177/0739532915600744}.
  \DOIprefix\doi{10.1177/0739532915600744}.
%Type = Book
\bibitem[{Risam(2018)}]{risam_new_2018}
\bibinfo{author}{R.~Risam}, \bibinfo{title}{New digital worlds: postcolonial
  digital humanities in theory, praxis, and pedagogy},
  \bibinfo{publisher}{Northwestern University Press},
  \bibinfo{address}{Evanston, Illinois}, \bibinfo{year}{2018}.
%Type = Book
\bibitem[{Noble(2018)}]{noble_2018}
\bibinfo{author}{S.~U. Noble}, \bibinfo{title}{Algorithms of Oppression: How
  Search Engines Reinforce Racism}, \bibinfo{publisher}{NYU Press},
  \bibinfo{year}{2018}. \URLprefix
  \url{http://www.jstor.org/stable/j.ctt1pwt9w5}.
%Type = Misc
\bibitem[{Labs(2017)}]{beyond_words}
\bibinfo{author}{L.~Labs}, \bibinfo{title}{Beyond words (``mark'')},
  \bibinfo{year}{2017}. \URLprefix
  \url{http://beyondwords.labs.loc.gov/\#/mark}.
%Type = Article
\bibitem[{Latham and Scholes(2006)}]{latham_scholes_2006}
\bibinfo{author}{S.~Latham}, \bibinfo{author}{R.~Scholes},
\newblock \bibinfo{title}{The rise of periodical studies},
\newblock \bibinfo{journal}{PMLA/Publications of the Modern Language
  Association of America} \bibinfo{volume}{121} (\bibinfo{year}{2006})
  \bibinfo{pages}{517–531}. \DOIprefix\doi{10.1632/003081206X129693}.
%Type = Misc
\bibitem[{con(NA)}]{conser_about_loc}
\bibinfo{title}{About {CONSER} - {Program} for {Cooperative} {Cataloging}
  ({Library} of {Congress})}, \bibinfo{year}{N/A}. \URLprefix
  \url{https://www.loc.gov/aba/pcc/conser/about/}.
%Type = Misc
\bibitem[{Ortiz~Baco(2019)}]{ortiz_baco_title_2019}
\bibinfo{author}{J.~Ortiz~Baco}, \bibinfo{title}{Title {Essays}, {Linked}
  {Data}, and the {Ethnic} {Press} in {Chronicling} {America}},
  \bibinfo{year}{2019}. \URLprefix
  \url{https://www.neh.gov/blog/title-essays-linked-data-and-ethnic-press-chronicling-america}.
%Type = Article
\bibitem[{Costanza-Chock and Rey-Maz{\'o}n(2016)}]{pageonex}
\bibinfo{author}{S.~Costanza-Chock}, \bibinfo{author}{P.~Rey-Maz{\'o}n},
\newblock \bibinfo{title}{Pageonex: New approaches to newspaper front page
  analysis},
\newblock \bibinfo{journal}{International Journal of Communication}
  \bibinfo{volume}{10} (\bibinfo{year}{2016}) \bibinfo{pages}{28}.
%Type = Article
\bibitem[{van~der Maaten and Hinton(2008)}]{tsne}
\bibinfo{author}{L.~van~der Maaten}, \bibinfo{author}{G.~Hinton},
\newblock \bibinfo{title}{Visualizing data using t-sne},
\newblock \bibinfo{journal}{Journal of Machine Learning Research}
  \bibinfo{volume}{9} (\bibinfo{year}{2008}) \bibinfo{pages}{2579--2605}.
  \URLprefix \url{http://jmlr.org/papers/v9/vandermaaten08a.html}.
%Type = Misc
\bibitem[{of~Congress(1891{\natexlab{a}})}]{appeal_14_march_1891}
\bibinfo{author}{L.~of~Congress}, \bibinfo{title}{The appeal, 14 march 1891.
  chronicling america: Historic american newspapers.},
  \bibinfo{year}{1891}{\natexlab{a}}. \URLprefix
  \url{https://chroniclingamerica.loc.gov/lccn/sn83016810/1891-03-14/ed-1/seq-1/}.
%Type = Misc
\bibitem[{of~Congress(1891{\natexlab{b}})}]{appeal_9_may_1891}
\bibinfo{author}{L.~of~Congress}, \bibinfo{title}{The appeal, 9 may 1891.
  chronicling america: Historic american newspapers.},
  \bibinfo{year}{1891}{\natexlab{b}}. \URLprefix
  \url{https://chroniclingamerica.loc.gov/lccn/sn83016810/1891-05-09/ed-1/seq-1/}.
%Type = Misc
\bibitem[{of~Congress(1896{\natexlab{a}})}]{bee_29_feb_1896}
\bibinfo{author}{L.~of~Congress}, \bibinfo{title}{The washington bee, 29 feb.
  1896. chronicling america: Historic american newspapers.},
  \bibinfo{year}{1896}{\natexlab{a}}. \URLprefix
  \url{https://chroniclingamerica.loc.gov/lccn/sn84025891/1896-02-29/ed-1/seq-1/}.
%Type = Misc
\bibitem[{of~Congress(1896{\natexlab{b}})}]{bee_29_aug_1896}
\bibinfo{author}{L.~of~Congress}, \bibinfo{title}{The washington bee, 29 aug.
  1896. chronicling america: Historic american newspapers.},
  \bibinfo{year}{1896}{\natexlab{b}}. \URLprefix
  \url{https://chroniclingamerica.loc.gov/lccn/sn84025891/1896-08-29/ed-1/seq-1/}.
%Type = Misc
\bibitem[{of~Congress(1900{\natexlab{a}})}]{ca_29_feb_1900}
\bibinfo{author}{L.~of~Congress}, \bibinfo{title}{The colored american, 7 april
  1900. chronicling america: Historic american newspapers.},
  \bibinfo{year}{1900}{\natexlab{a}}. \URLprefix
  \url{https://chroniclingamerica.loc.gov/lccn/sn83027091/1900-04-07/ed-1/seq-1/}.
%Type = Misc
\bibitem[{of~Congress(1900{\natexlab{b}})}]{ca_8_sep_1900}
\bibinfo{author}{L.~of~Congress}, \bibinfo{title}{The colored american, 8 sept.
  1896. chronicling america: Historic american newspapers.},
  \bibinfo{year}{1900}{\natexlab{b}}. \URLprefix
  \url{https://chroniclingamerica.loc.gov/lccn/sn83027091/1900-09-08/ed-1/seq-1/}.

\end{thebibliography}

\end{document}